\documentclass{article} %
\usepackage{iclr2024_conference,times}

\usepackage{amsmath,amsfonts,bm}

\def\eqref#1{equation~\ref{#1}}

\def\1{\bm{1}}

\DeclareMathAlphabet{\mathsfit}{\encodingdefault}{\sfdefault}{m}{sl}
\SetMathAlphabet{\mathsfit}{bold}{\encodingdefault}{\sfdefault}{bx}{n}

\usepackage{natbib}
\usepackage[utf8]{inputenc} %
\usepackage[T1]{fontenc}    %
\usepackage{url}            %
\usepackage{booktabs}       %
\usepackage{amsfonts}       %
\usepackage{nicefrac}       %
\usepackage{microtype}      %
\usepackage{xcolor}         %
\usepackage{amsmath}
\usepackage{todonotes}
\usepackage{mathabx}
\usepackage{amssymb} 
\usepackage{multirow}
\usepackage{caption}
\usepackage{authblk}
\usepackage[misc]{ifsym} 
\usepackage{subcaption}
\usepackage{color}
\usepackage{colortbl}

\definecolor{linkc}{rgb}{0, 0.44, 0.74}
\definecolor{eqc}{rgb}{1, 0, 0}
\usepackage[pagebackref=false,breaklinks=true,colorlinks=True,urlcolor=eqc,citecolor=linkc,linkcolor=eqc,bookmarks=false]{hyperref}

\def\x{{\bm{x}}}
\def\c{{\bm{c}}}

\def\I{{\bm{I}}}
\newcommand{\mymid}{\,|\,}

\newcommand{\blfootnote}[1]{\begingroup
\renewcommand\thefootnote{}\footnote{#1}\addtocounter{footnote}{-1}
\endgroup}
\title{Image Conductor: 
Precision Control for Interactive Video Synthesis}

\author{
  \begin{minipage}[t]{\textwidth}
    \raggedright
    \textbf{Yaowei Li$^{1,2}$, \space\space\space 
    Xintao Wang$^{2}$, \space\space\space
    Zhaoyang Zhang$^{2\dagger}$, \space\space\space
    Zhouxia Wang$^{2,3}$}, \\ \vspace{-3mm}
    \textbf{Ziyang Yuan$^{2,4}$, \space\space\space
    Liangbin Xie$^{2,5,6}$, \space\space\space
    Yuexian Zou$^{1\text{\Letter}}$, \space\space\space
    Ying Shan$^{2}$ }
    \\
    \vspace{1mm}
    $^{1}$\normalfont{Peking University}\space
    $^{2}$\normalfont{ARC Lab, Tencent PCG} \space
    $^{3}$\normalfont{Nanyang Technological University} \\
    $^{4}$\normalfont{Tsinghua University} \space
    $^{5}$\normalfont{University of Macau} \space
    $^{6}$\normalfont{Shenzhen Institute of Advanced Technology} 
    \\
    \vspace{1mm}
    Project Page: \url{https://liyaowei-stu.github.io/project/ImageConductor/}
  \end{minipage}
}

\iclrfinalcopy %

\begin{document}

\makeatletter
\let\@oldmaketitle\@maketitle
\renewcommand{\@maketitle}{
\@oldmaketitle
\vspace{-12mm}
\begin{center} %
\begin{minipage}{0.8\linewidth}
\centering
\includegraphics[width=\textwidth]{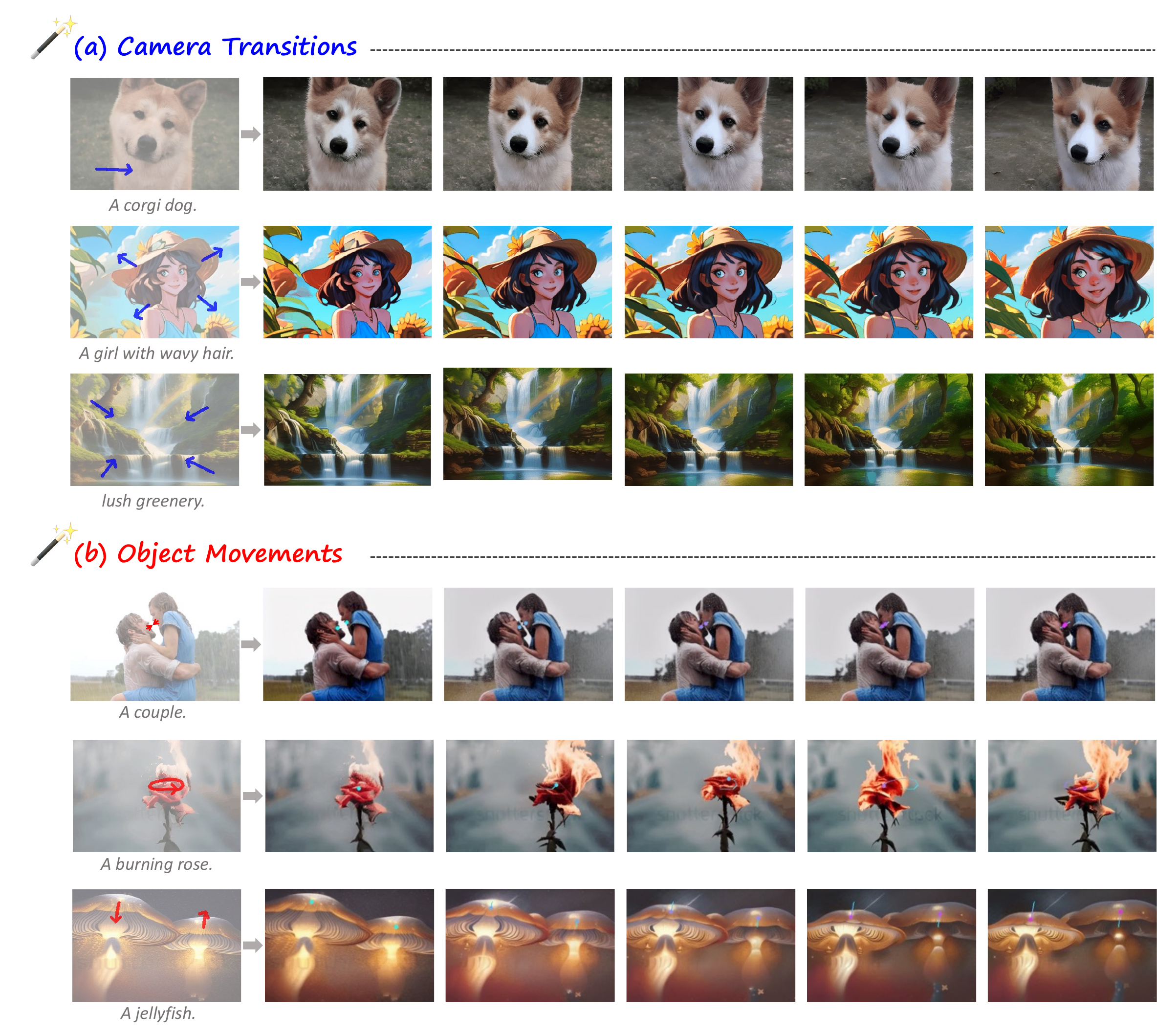}
\captionof{figure}{\textbf{Orchestrated Results of Image Conductor.} Image Conductor enables fine-grained and accurate image-to-video motion control, including both camera transitions and object movements. Colorful lines denote motion trajectories. \label{fig:teaser}}
\vspace{1mm}
\end{minipage}
\end{center}
  }
  
\makeatother

\maketitle

\blfootnote{\Letter \space Corresponding author. $\dagger$ Project lead.}

\vspace{-5mm}
\begin{abstract}
Filmmaking and animation production often require sophisticated techniques for coordinating camera transitions and object movements, typically involving labor-intensive real-world capturing.
Despite advancements in generative AI for video creation, achieving precise control over motion for interactive video asset generation remains challenging. 
To this end, we propose Image Conductor, a method for precise control of camera transitions and object movements to generate video assets from a single image.
An well-cultivated training strategy is proposed to separate distinct camera and object motion by camera LoRA weights and object LoRA weights. 
To further address cinematographic variations from ill-posed trajectories, we introduce a camera-free guidance technique during inference, enhancing object movements while eliminating camera transitions.
Additionally, we develop a trajectory-oriented video motion data curation pipeline for training. 
Quantitative and qualitative experiments demonstrate our method's precision and fine-grained control in generating motion-controllable videos from images, advancing the practical application of interactive video synthesis.
\end{abstract}

\section{Introduction}

\vspace{-2mm}
Filmmaking and animation production are essential forms of visual art.
During the creative process of video media, professional directors often require advanced cinematography techniques to meticulously plan and coordinate camera transitions and object movements, ensuring storyline coherence and refined visual effects. 
To achieve precise creative expression, the current workflow for video media orchestration and production heavily relies on real-world capturing and 3D scan modeling, which are labor-intensive and costly.

Recent work~\citep{ho2022imagen, blattmann2023align, girdhar2023emu, xing2023dynamicrafter, chen2023videocrafter1, blattmann2023stable, bar2024lumiere, videoworldsimulators2024} explores an AIGC-based filmmaking pipeline that leverages the powerful generative capabilities of diffusion models to  generate video clip assets.
Despite these advancements, generating dynamic video assets allowing   creators precisely  express  their ideas remains unusable, for:
(1) Lacking of efficient  generating control interface.
(2) Lacking of fine-grained and accurate control over camera transitions and object movements.

Although several works have attempted to introduce motion control signals to guide the video generation process~\citep{yin2023dragnuwa, wang2023motionctrl, wang2024boximator, wu2024draganything}, none of the existing methods support accurate and fine-grained control over both camera transitions and object movements (see Fig.~\ref{fig:teaser}).

In fact, data available on the internet often mixes both camera transitions and object movements, leading to ambiguities between the two types of motion. Although MotionCtrl~\citep{wang2023motionctrl} uses a data-driven approach to decouple camera transitions from object motion, it still lacks precision and effectiveness. Camera parameters are neither intuitive nor straightforward to obtain for cinematographic variations. For object movements, MotionCtrl uses ParticleSfM~\citep{zhao2022particlesfm}, a motion segmentation network based on optical flow estimation, which introduces significant errors. Additionally, ground truth videos annotated based on motion segmentation networks still contain camera transitions, causing generated videos to exhibit unintended cinematographic variations.
Decoupling cinematographic variations from object movements through data curation is inherently challenging. Obtaining video data from a fixed camera viewpoint, i.e., videos with only object movements, is difficult. Optical flow-based motion segmentation methods~\citep{teed2020raft, xu2022gmflow, zhao2022particlesfm, yin2023dragnuwa, wang2023motionctrl} struggle to accurately track moving objects without errors and fail to eliminate intrinsic camera transitions in realistic videos. Overall, existing methods are either not fine-grained or not sufficiently effective.

In this paper, we propose Image Conductor, an interactive method for fine-grained object motion and camera control to generate accurate video assets from a single image. Effective fine-grained motion control requires robust motion representation. 
Trajectories, being intuitive and user-friendly, allow users to control motion in video content by drawing paths. However, a large-scale, high-quality open-source trajectory-based tracking video dataset is currently lacking. To address this, we use CoTracker~\citep{karaev2023cotracker} to annotate existing video data and design a data filtering workflow, resulting in high-quality trajectory-oriented video motion data.

To address the coupling of cinematographic variations and object movements in real-world data, we first train a video ControlNet~\citep{zhang2023adding} using annotated data to convey motion information to the UNet backbone of the diffusion model. We then propose a collaborative optimization method that applies distinct sets of Low-Rank Adaptation (LoRA) weights~\citep{hu2021lora} on the ControlNet to distinguish various types of motion. 
In addition to the denoising loss commonly used in diffusion models, we introduce an orthogonal loss to ensure the independence of different LoRA weights, enabling accurate motion disentanglement.

To flexibly eliminate cinematographic variations caused by ill-posed trajectories, which are difficult to distinguish in LoRA, and to enhance object movement, we also introduce a new camera-free guidance. This technique iteratively executes an extrapolation fusion between different latents during the sampling process of diffusion models, similar to the classifier-free guidance technique~\citep{ho2022classifier}.

In brief, our main contributions are as follows:
\begin{itemize}
\item We construct a high-quality video motion dataset with precise trajectory annotations, addressing the lack of such data in the open-source community.
\item We introduce a method to collaboratively optimize LoRA weights in motion ControlNet, effectively separating and controlling camera transitions and object movements
\item We propose camera-free guidance to heuristically eliminate camera transitions caused by multiple trajectories that are challenging to separate with LoRA weights.
\item Extensive experiments demonstrate the superiority of our method in precisely and finely motion control, enabling the generation of videos from images that align with user desires.
\end{itemize}

\section{Approach}
\subsection{Overview}
Image Conductor aims to animate a static image by precisely directing camera transitions and object movements according to user specifications, producing coherent video assets. Our workflow includes trajectory-oriented video data construction (Sec.~\ref{Sec:3.2}), a motion-aware image-to-video architecture (Sec.~\ref{Sec:3.3}), controllable motion separation (Sec.~\ref{Sec:3.4}), and camera-free guidance (Sec.~\ref{Sec:3.5}).

We use user-friendly trajectories to define the intensity and direction of camera transitions and object movements. 
To address the lack of large-scale annotated video data, we design a data construction pipeline to create a consistent video dataset with appropriate motion.

Using this data, we train Video ControNet~\citep{zhang2023adding} to synthesize motion-controllable video content. To eliminate ambiguities between camera transitions and object movements, we employ separate sets of LoRA weights. 
First, we train with camera-only LoRA weights to control camera transitions. Then, we load these weights and use a new set of object LoRA weights to decouple object movement, ensuring precise control. We also introduce a loss function with orthogonal constraints to maintain independence between different LoRA weights.

\begin{figure}
\centering
\includegraphics[width=\textwidth]{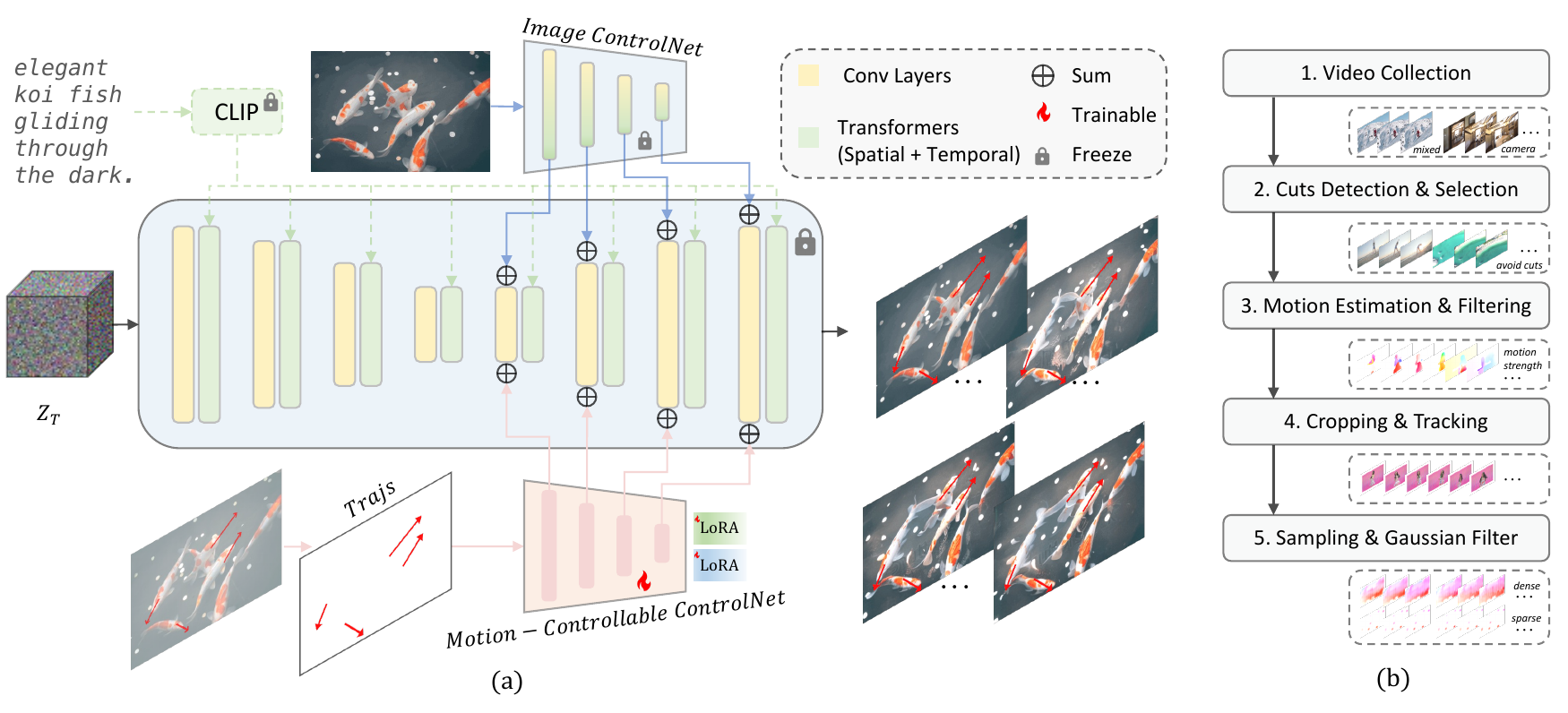}
\caption{\textbf{ a) Framework of Image Conductor.} 3D UNet serves as the diffusion backbone, while image ControlNet and motion-controllable ControlNet (and its LoRA weights) convey appearance and motion information, respectively. We progressively fine-tune different modules during trarning phase (see Sec~\ref{Sec:3.4}).   \textbf{b) Trajectory-oriented video motion data construction workflow.} We carefully curate the data to ensure dynamic and consistent video content, as well as precise trajectory annotations (see Sec~\ref{Sec:3.2}). \label{fig:_framework}}
\end{figure}

\begin{figure}
\centering
\includegraphics[width=\textwidth]{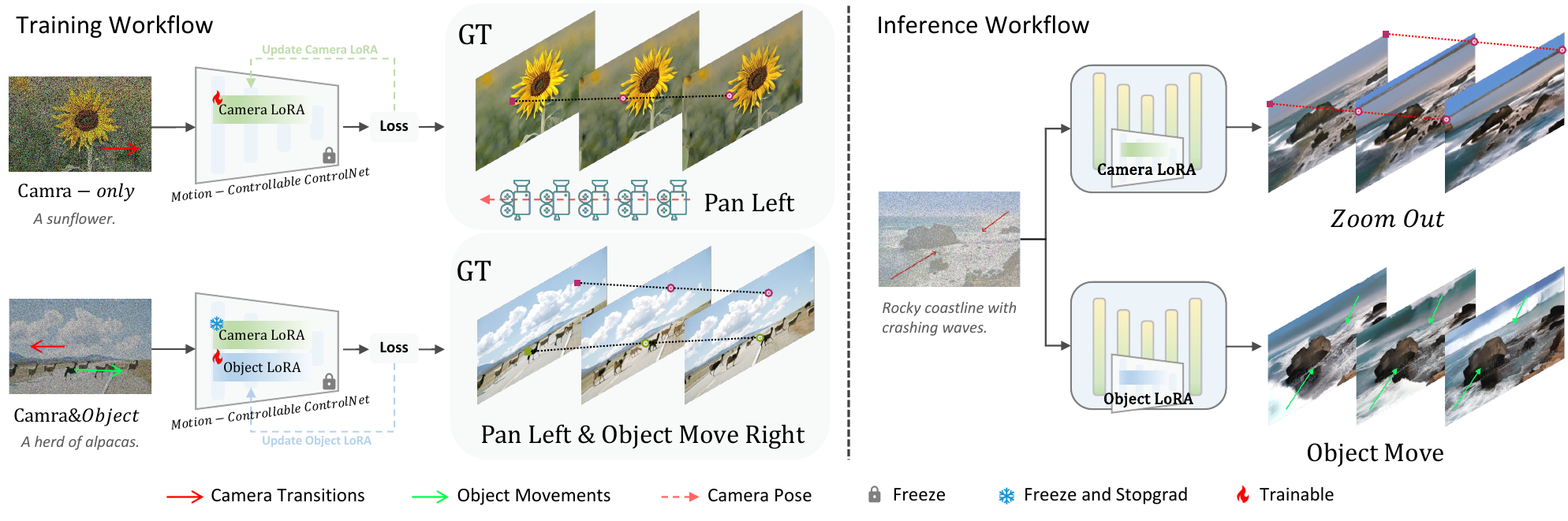}
\caption{\textbf{Fine-grained Motion Separation Method.} a) The training process is divided into two stages. Initially, camera-only data is used to empower the camera LoRA with the ability to control camera transitions.  After loading the well-trained camera LoRA, mixed motion data is used to train the object LoRA, refining object motion information. b) During inference, loading different LoRAs provides the model with various control capabilities. \label{fig:_idea} }
\end{figure}

To seamlessly blend camera transitions and object movements, we propose a camera-free guidance technique that iteratively extrapolates between camera and object motion latents during inference. 
Fig.~\ref{fig:_framework} (a) shows our framework, Fig.~\ref{fig:_framework} (b) illustrates our data curation pipeline, and Fig.~\ref{fig:_idea} presents the core idea of Image Conductor.

\subsection{Trajectory-Oriented Video Motion Data Construction.}
\label{Sec:3.2}
Since Image Conductor relies on trajectories to guide motion, we need a dataset with trajectory annotations to track dynamic information in videos. Existing large-scale video datasets typically lack such annotations. While some methods use motion estimators to annotate video data, these approaches often suffer from inaccuracies\citep{yin2023dragnuwa, wang2023motionctrl, wu2024draganything} or lack generality\citep{wu2024draganything}. 
Moreover, almost all annotated datasets with trajectory annotations are not publicly available. To address this, we introduce a comprehensive and general pipeline for generating high-quality video data with appropriate motion and consistent scenes, as illustrated in Fig.~\ref{fig:_framework} (b).

\paragraph{Video Collection.}
We utilize the WebVid dataset\citep{bain2021frozen}, a large-scale mixed dataset with textual descriptions, and the Realestate10K dataset\citep{zhou2018stereo}, a camera-only dataset, for our research. The Image Conductor aims to decouple object movements from mixed data, requiring scene consistency and high motion quality. 
To ensure temporal quality, we process the WebVid dataset by detecting cuts and filtering motion. For the Realestate10K dataset, we focus on the diversity of camera transitions and generate video captions using BLIP2\citep{blip2} by extracting frames at specific intervals and concatenating their descriptions.

\paragraph{Cuts Detection and Selection.}
In videos, cuts refer to transitions between different shots, and generative video models are sensitive to such motion inconsistencies~\citep{blattmann2023stable}. To avoid cuts and abrupt scene changes, which can cause the model to overfit these phenomena, we first use a cut detection tool~\protect\footnotemark{} to identify cuts within the video dataset. We then select the longest consistent scenes as our video clips, ensuring scene consistency.

\footnotetext{\url{https://github.com/Breakthrough/PySceneDetect}.}

\paragraph{Motion Estimation and Filtering.}
o ensure the dataset exhibits good dynamics, we use RAFT~\citep{teed2020raft} to compute the optical flow between adjacent frames and calculate the Frobenius norm as a motion score. We filter out the lowest 25\% of video samples based on this score. To reduce computational cost, we resize the shorter side of the videos to 256 pixels and randomly sample a 32-frame sequence with a temporal interval of 1 to 16 frames. These 32 frames are used as the training dataset, and their motion scores are computed for sample filtering.

\paragraph{Cropping and Tracking.}
To standardize the dimensions of the training data, we perform center cropping on the previously obtained data, resulting in video frames of size $384\times256\times32$. We then employ CoTracker~\citep{karaev2023cotracker}, a tracking method towards dense point, to record motion within the video using a $16\times16$ grid. 
Compared to optical flow-based point correspondence methods~\citep{teed2020raft, xu2022gmflow}, tracking avoids drift-induced error accumulation, providing a more accurate representation of motion. 
After tracking, we accumulate point trajectories by calculating the differences between adjacent points within the same trajectory. This results in stacked flow maps compatible with the input format of ControlNet~\citep{zhang2023adding}.

\paragraph{Sampling and Gaussian Filter.}
o enhance user interaction and usability, we use sparse trajectories for motion guidance. We heuristically sample   $n\in[1,8]$  trajectories from the dense set, with 8 being the upper limit. The value of  $n$ is randomly selected, and the normalized motion intensity of each trajectory is used as the sampling probability. The accumulated flow map from these trajectories forms a sparse matrix. 
To avoid training instability caused by the sparse matrix, we apply a Gaussian filter to the trajectories, similar to previous methods~\citep{yin2023dragnuwa, wang2023motionctrl, wu2024draganything}. Through this data processing pipeline, we constructed a trajectory-oriented video motion dataset containing 
130k mixed videos with camera transitions and object movements, and 
62k videos with only camera transitions.

\subsection{Motion-aware Image-to-video Architecture}
\label{Sec:3.3}

\paragraph{Image-to-Video Backbone.} 
As illustrated in Fig.~\ref{fig:_framework} (a), we utilize Animatediff~\citep{guo2023animatediff} equipped with SparseCtrl~\citep{guo2023sparsectrl} for images as our pre-trained image-to-video foundational model. 
This model uses the CLIP~\citep{clip} text encoder to extract text embeddings $c_{txt}\in\mathbb{R}^{1\times d}$, which are then passed to the UNet~\citep{UNet} backbone via cross-attention mechanism. 
The input image, serving as the first frame, is concatenated with an all-zero frame matrix and a mask identifier channel-wise to form $c_{img}\in\mathbb{R}^{T \times 4 \times H \times W}$. 
Next, the video SparseCtrl, a variant of the ControlNet~\citep{ControlNet} that removes the skip-connections between the ControlNet's and the UNet encoder's input latents, is used to extracts image information from $c_{img}$.

\paragraph{Motion-Controllable ControlNet.}
To extract motion information from the annotated trajectory input $c_{trajs}\in\mathbb{R}^{T \times 2 \times H \times W}$ for composition of camera transitions and object movements in videos, we use ControlNet as the motion encoder to capture multi-level motion representations. 
This ControlNet incorporates different types of LoRA weights to guide the image-to-video generation with user-desired camera transitions and object movements. Consistent with the observations of SparseCtrl~\citep{guo2023sparsectrl}, we find that removing the skip connections between the main branch's and the conditional branch's input latents speeds up convergence during training.

\subsection{Controllable Motion Separation}
\label{Sec:3.4}
The aim of our approach is to precisely separate camera transitions and object movements in videos, enabling fine-grained control over the generation of video clip asserts that meets user expectations. To this end, we introduced camera LoRA  $\theta_\text{cam}$  and object LoRA $\theta_\text{obj}$  into the motion ControlNet to guide the synthesis of different types of motion. 
As shown in the Fig.~\ref{fig:_idea}, during the training process, we employed a collaborative optimization strategy. First, we optimized the camera LoRA, and then, we optimized the object LoRA based on the loaded camera LoRA. During the inference stage, the model loads different LoRA to control camera transitions (e.g., zooming out) and object movements (e.g., two waves advancing in a specified direction).

\paragraph{Camera Transitions.}
Since it is available to obtain data with camera-only transition, we straightforwardly train camera LoRA $\theta_\text{cam} = \theta_0 + \Delta\theta_\text{cam}$ using our our carefully cultivated camera motion dataset, endowing the ControNet with the ability to direct cinematographic variations. The standard diffusion denoising training objective is utilized:
\begin{equation}
  \mathcal{L_\text{cam}} = \mathbb{E}_{z_{0,\text{cam}}, c_\text{txt}, c_\text{img}, c_\text{trajs}, \epsilon\sim\mathcal{N}(0, \mathit{I}), t} \left\lbrack
  \lVert \epsilon - \epsilon_{\theta_\text{cam}(z_{t,\text{cam}}}, t, c_\text{txt}, c_\text{img}, c_\text{trajs}) \rVert_2^2 \right\rbrack,
  \label{euqation: noise_preditcion_loss_cam}
\end{equation}
where $\theta_\text{cam}$ is the denoiser where ControlNet with camera LoRA loaded,  $z_{t,\text{cam}}$ is the noisy latent of videos with only camera transition at timestep $t$, $c_\text{txt}$, $c_\text{img}$ and $c_\text{trajs}$ refer to the text prompt, image prompt, and conditional trajectory, respectively.

\paragraph{Object Movements.}
Due to the scarcity of fixed-camera-view video data without cinematographic variations, we need to decouple object motion from mixed data where both camera transitions and object movements are exsist. 
Observing that distinct types of motion share the same trajectory, we can further train the object LoRA $\theta_{\text{obj}} = \theta_0 + \Delta\theta_\text{obj}$ after loading the well-trained camera LoRA weights, i.e., targeting the reconstruction of camera transitions and object movements in the original video content from mixed data. Formally, we load both the camera LoRA and object LoRA simultaneously during training phase, and prevent gradient flow to the camera LoRA via stopgrad $\text{sg}[\cdot]$: 
\begin{equation}
    \label{eq:mixed}
    \begin{aligned}
        \theta_{\text{mixed}} = \theta_0 + \text{sg}[\Delta \theta_{\text{cam}}] + \Delta \theta_{\text{obj}}.
    \end{aligned}
\end{equation}

Similarly, we optimize the object LoRA using the standard diffusion denoising objective:
\begin{equation}
  \mathcal{L_\text{mixed}} = \mathbb{E}_{z_{0,\text{mixed}}, c_\text{txt}, c_\text{img}, c_\text{trajs}, \epsilon\sim\mathcal{N}(0, \mathit{I}), t}\left\lbrack
  \lVert \epsilon - \epsilon_{\theta_\text{mixed}(z_{t,\text{mixed}}}, t, c_\text{txt}, c_\text{img}, c_\text{trajs}) \rVert_2^2 \right\rbrack,
  \label{euqation: noise_preditcion_loss_obj}
\end{equation}
where $\theta_\text{mixed}$ is the denoiser where ControlNet with all LoRA loaded as in Eq.~\ref{eq:mixed},  $z_{t,\text{cam}}$ is the noisy latent of videos with camera transition and object movements at timestep $t$.

\paragraph{Orthogonal Loss.}
To encourage the object LoRA to learn concepts distinct from the camera LoRA and to accelerate the convergence of the model, we propose an orthogonal loss as a joint optimization objective. Specifically, we extract all linear layer weights $W_{\text{cam}}$ and $W_{\text{traj}}$ from the different LoRAs and impose an orthogonality constraint on them:
\begin{equation}
  \mathcal{L_\text{ortho}} = \mathbb{E}_{W_{i,\text{cam}}\in W_{\text{cam}}, W_{i,\text{traj}}\in W_{\text{traj}}}\left\lbrack
  \lVert{I - W_{i,\text{cam}} W_{i,\text{traj}}^T}\rVert_2^2 \right\rbrack
  \label{euqation: ortho_loss}
\end{equation}
where $I$ represents the identity matrix, $W_{i,\text{cam}}$ and $W_{i,\text{traj}}$ refer to  the weights of the $i$-th linear layer of the camera LoRA and object LoRA, respectively.

In all, the optimization process is incremental. We first optimize the camera LoRA using $\mathcal{L_\text{cam}}$, and then optimize the object LoRA using $\mathcal{L_\text{mixed}}$ and $\mathcal{L_\text{ortho}}$.

\subsection{Camera-free Guidance}
\label{Sec:3.5}
When users aim to control multiple objects, multiple trajectories often introduce camera transitions. Inspired by classifier-free guidance~\citep{ho2022classifier}, we propose a camera-free guidance technique to flexibly and seamlessly enhance motion intensity while eliminating camera transitions. 
\begin{equation}
\label{eq:cfg}
    \begin{aligned}
         \hat{\epsilon}_{\theta_0,\theta_\text{trajs}}(\x_t, \c) = & \epsilon_{\theta_0}(\x_t, \varnothing) \\
        +\ & \lambda_{\text{cfg}} (\epsilon_{\theta_0}(\x_t, \c) - \epsilon_{\theta_0}(\x_t, \varnothing)) \\
       +\ & \lambda_{\text{trajs}}(\epsilon_{\theta_{\text{trajs}}}(\x_t, \c) - \epsilon_{\theta_0}(\x_t, \c)),
    \end{aligned}
\end{equation}
where $\theta_\text{trajs}$ refer to the model with object LoRA and $\theta_0$ is the model with pre-trained motion ControlNet. The final output latent is derived by extrapolating the outputs of these two components.

\section{Experiments}

\subsection{Comparisons with State-of-the-Art Methods}
We compare Image Conductor with exsisting state-of-the-art image-based or text-based motion controllable video generation methods, namely DragNUWA~\citep{yin2023dragnuwa}, DragAnything~\citep{wu2024draganything} and MotionCtrl~\citep{wang2023motionctrl}.

\paragraph{Evaluation datasets.}
To independently evaluate camera transitions and object movements, we use two distinct datasets:
1) Camera-Only Motion Evaluation Dataset: We select 10 camera trajectories, \textit{e.g.} pan left, pan right, pan up, pan down, zoom in, zoom out, to evaluate control over cinematographic variations.
2) Object-Only Motion Evaluation Dataset: We design 10 varied trajectories, including straight lines, curves, shaking lines, and their combinations. 

\begin{figure}[ht!]
\centering
\includegraphics[width=.9\textwidth]{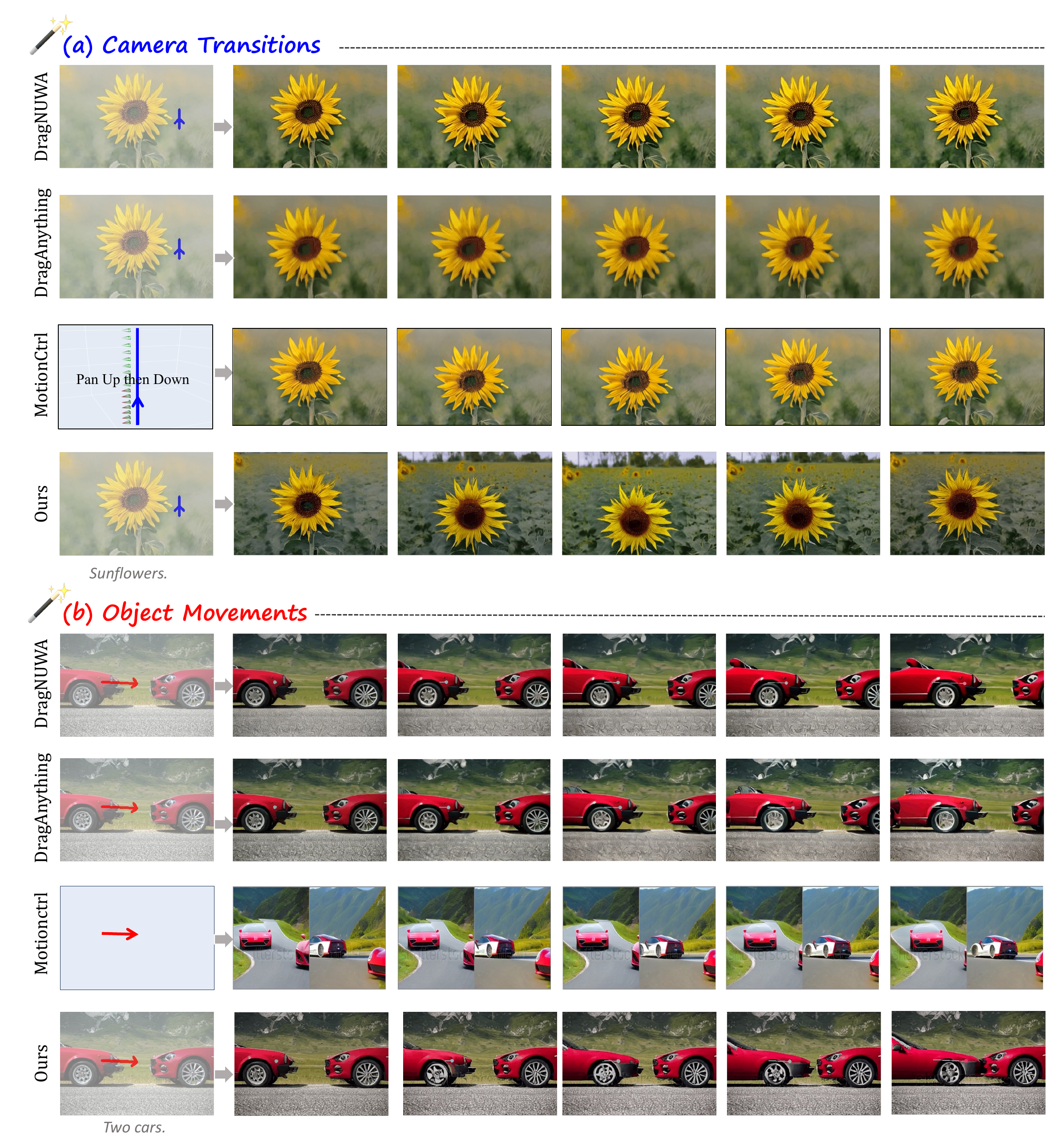}
\caption{\textbf{Qualitative Comparisons of the proposed Image Conductor.} (a) Camera Transitions. Our method can simultaneously utilize text, image, and trajectory prompts as control signals to achieve more natural content and camera transitions. (b) Object Movements. Apart from our method, other approaches incorrectly confuse object movements with camera transitions. \label{fig:_comparision}}
\vspace{-10pt}
\end{figure}

\paragraph{Qualitative Evaluation}
Fig.~\ref{fig:_comparision} displays some of our qualitative results. Compared to previous methods~\citep{yin2023dragnuwa, wu2024draganything, wang2023motionctrl}, our approach can effectively control camera transitions and object movements. In terms of camera transitions, both DragNUWA and DragAnything fail to achieve the camera transition of panning down and then up in the generated video. Although Motionctrl-SVD is capable of generating the specified camera movement, it is unable to define natural content changes via text prompts. Additionally, it cannot accurately define the intensity of camera changes, and sometimes introduces distortion artifact. 

In terms of object movements, both DragNUWA and DragAnything incorrectly interpret object movement as camera transition, resulting in generated videos that do not meet user intentions. In addition, the motion trajectories of their generated videos are often poorly matched to the desired trajectories precisely due to the errors introduced by the labeled dataset. As trajectory-based MotionCtrl relies on the text-to-video model, we directly use text and trajectory prompts to control the generation of the video under different seed. The results demonstrate that it lacks fine-grained control over the generated content due to its inability to use images as conditions. Additionally, it still exhibits a significant amount of camera transition rather than object movement. In all, our method is capable of accurately and finely controlling various types of motion utilizing the separated LoRA.

\begin{figure}[htp]
\centering
\includegraphics[width=\textwidth]{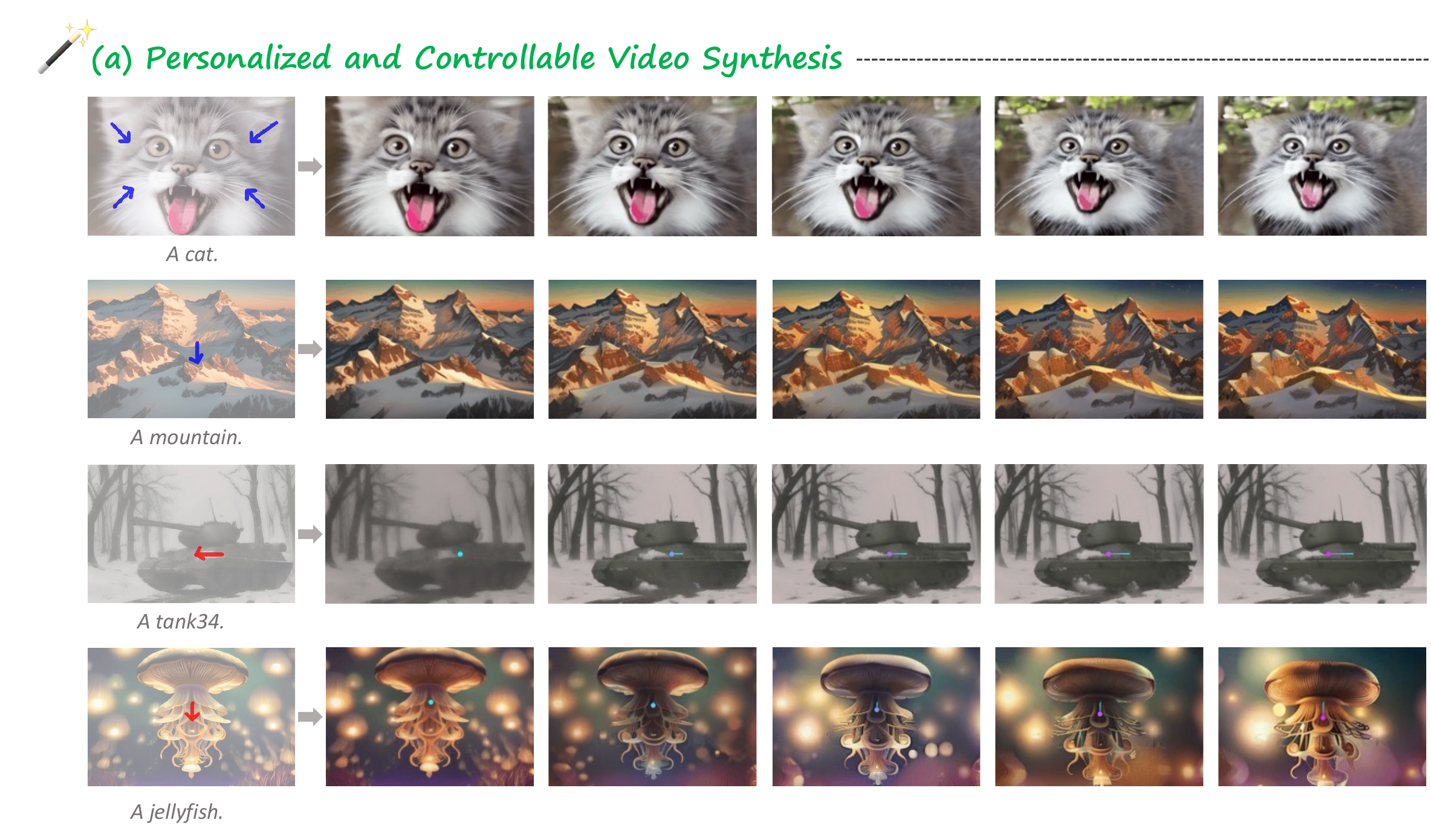}
\caption{\textbf{Results of Personalized and Controllable Video Synthesis.} The pre-trained base model and LoRA weights are sourced from TuSun~\protect\footnotemark{}, HelloObject~\protect\footnotemark{}, and CardosAnime~\protect\footnotemark{} checkpoint.}
\label{fig:personalize}
\vspace{-5pt}
\end{figure}

\paragraph{Quantitative Evaluation}
As shown in the Tab.~\ref{tb_cp}, compared to other methods, our proposed Image Conductor achieves state-of-the-art quantitative performance. We measure our alignment with the given trajectoies via the CamMC and ObjMC metrics, surpassing the baseline models and demonstrating our precise motion control capabilities. At the same time, the FID and FVD metrics illustrate that our generation quality surpasses other models, capable of producing realistic videos.  Furthermore, we invite 31 participants to assess the results of DragNUWA, DragAnything and Image Conductor. The assessment includes video quality, motion similarity. Participants are also asked to give an overall preference for each compared pair. The statistical results confirm  that our generated videos not only appear more realistic and visually appealing but also exhibit superior motion adherence compared to those produced by other models.

\addtocounter{footnote}{-2}
\footnotetext{\url{https://civitai.com/models/33194/leosams-pallass-catmanul-lora}.}
\addtocounter{footnote}{1}
\footnotetext{\url{https://civitai.com/models/121716/helloobjects}.}
\addtocounter{footnote}{1}
\footnotetext{\url{https://civitai.com/models/25399/cardos-anime}.}

\begin{table*}[t]
\caption{\textbf{Quantitative Comparisons with SOTA Methods.} We utilize automatic metrics (\textit{i.e.}, FID, FVD, CamMC, ObjMC) and human evaluation (\textit{i.e.}, overall performance, sample quality, motion similarity) to evaluate the performance. DN and DA denotes DragNUWA~\citep{wu2024draganything} and DragAnything~\citep{yin2023dragnuwa}, respectively.}
\vspace{-5pt}
\small
\centering
\begin{tabular}{c | c c c c|c c c }
\toprule
\multirow{2}{*}{Method} & \multicolumn{4}{c|}{Automatic Metrics} & \multicolumn{3}{c}{Human Evaluation}\\
 & FID $\downarrow$ & FVD $\downarrow$ & CamMC $\downarrow$ & ObjMC $\downarrow$ & Overall $\uparrow$ & Quality $\uparrow$ & Motion $\uparrow$\\
\hline
DN~\citep{yin2023dragnuwa} & 237.26 & 1283.85 & 48.72 & 51.24 & $31.8\%$ & $37.1\%$ & $27.7\%$\\
DA~\citep{wu2024draganything} & 243.17 & 1287.15 & 66.54 & 60.97 & $6.5\%$ & $8.1\%$ & $6.3\%$\\
Image Conductor & \textbf{209.74} & \textbf{1116.17} & \textbf{33.49} & \textbf{42.38} & $\mathbf{61.7\%}$ & $\mathbf{54.8}$\% & $\mathbf{66.0}$\% \\
\toprule
\end{tabular}
\label{tb_cp}
\vspace{-10pt}
\end{table*}

\subsection{Personalized and Controllable Video Synthesis}
Since the base T2V model is not fine-tuned, our method naturally possesses the ability for personalized generation while maintaining controllability. In Fig.~\ref{fig:personalize}, we loaded some personalized models to sample videos using the provided prompt, guidance scale and user-specified trajectories. The results show that our method can seamlessly integrate with open-source customization communities (e.g., CIVITAI~\protect\footnotemark{}) and has powerful capabilities for generating controllable video content assets.

\footnotetext{\url{https://civitai.com/}.}

\begin{figure}[htp!]
\centering
\includegraphics[width=0.9\textwidth]{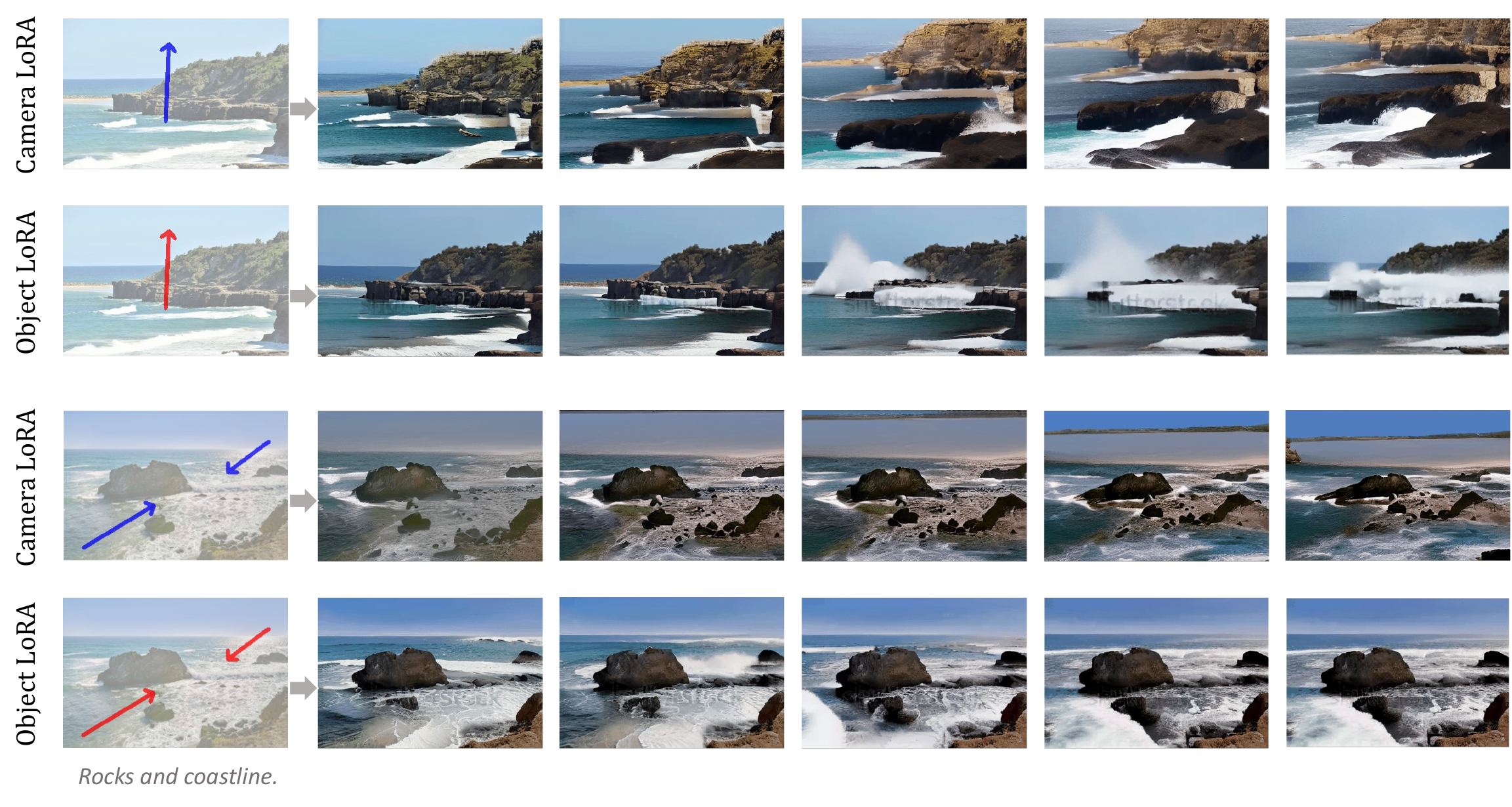}
\vspace{-5pt}
\caption{\textbf{Effect of distinct LoRA weights.}  Image conductor enables users to independently control camera and object motion interactively. \label{fig:abl1}}
\vspace{-5pt}
\end{figure}

\begin{figure}[h!]
\centering
\includegraphics[width=0.9\textwidth]{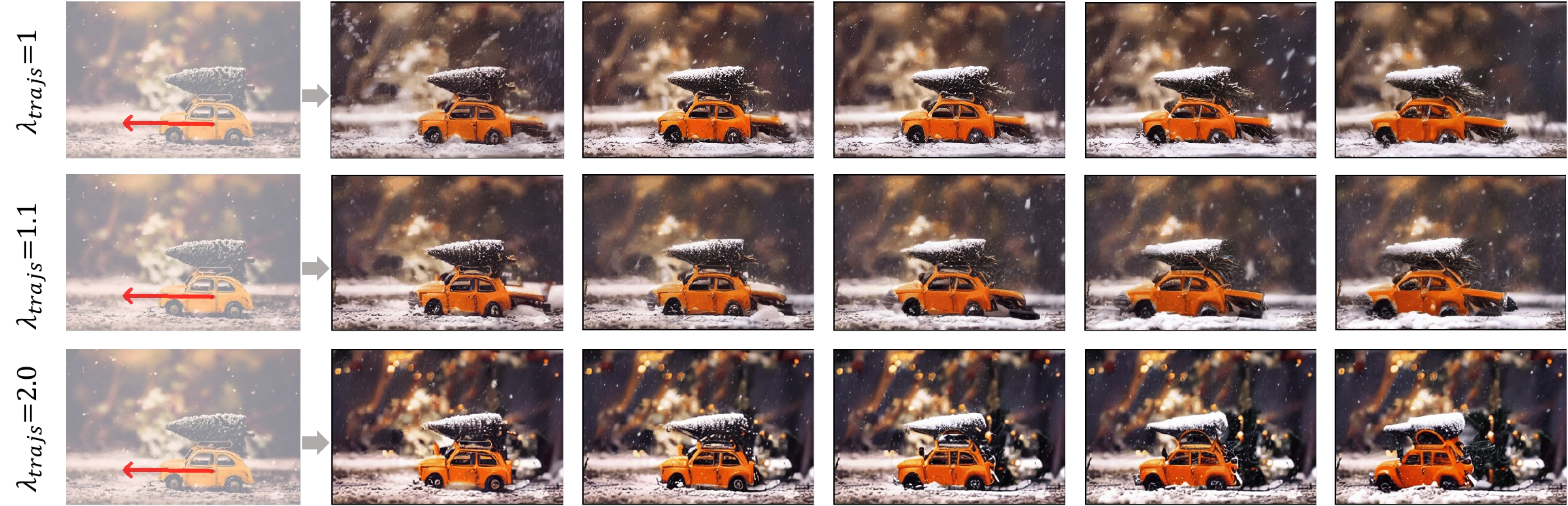}
\vspace{-5pt}
\caption{\textbf{Effect of Camera-free Guidance.} The camera-free guidance approach flexibly enhances object movements during inference.  \label{fig:abl2}}
\vspace{-5pt}
\end{figure}

\subsection{Ablation studies}

\paragraph{Effect of Distinct LoRA Weights}
To validate that our carefully designed interactive optimization strategy can separate camera transitions and object movements through distinct LoRA weights, we use the same trajectory as input to guide different LoRA to generate videos. As shown in Fig.~\ref{fig:abl1}, loading various LoRA weights endows the model with different capabilities. For instance, a vertically upward trajectory causes the video to pan up when the camera LoRA is loaded, and it creates upward waves when the object LoRA is loaded.

\paragraph{Effect of Camra-free Guidance}
As shown in Fig.~\ref{fig:abl2}, using camera-free guidance can facilitate the separation of object movements from camera transitions in several challenging examples. When  camera-free guidance $\lambda_{\text{trajs}}$ is set to 1, \textit{i.e}., camera-free guidance is not yet used, the generated video exhibits a unexpected pan left transformation. When the $\lambda_{\text{trajs}}$ is set to 1.1, the generated videos exhibit reasonable object movements, yet some artifacts still remain. As the guidance increases, the movements of the object becomes more apparent and clear.

\section{Conclusion}
In conclusion, this paper introduces Image Conductor, a novel approach for precise and fine-grained control of camera transitions and object movements in interactive video synthesis. 
We  design a training strategy and utilized distinct LoRA weights to decouple camera transition and object movements. 
Additionally, we propose a camera-free guidance technique to enhance object movement control. Extensive experiments demonstrate the effectiveness of our method, marking a significant step towards practical applications in video-centric creative expression.

\newpage

\bibliography{iclr2024_conference}
\bibliographystyle{iclr2024_conference}

\clearpage
\newpage
\appendix

\section{Related Works}
\paragraph{Video Synthesis.} With the emergence of massive data~\citep{bain2021frozen, chen2024panda} and the gradual perfection of diffusion model theory~\citep{ho2020denoising, song2020score, batzolis2021conditional}, deep generative models have made remarkable progress~\citep{ldm, TI, ControlNet, wang2023videocomposer, chen2023videocrafter1, videoworldsimulators2024}. Despite the significant achievements, current video generation methods~\citep{wang2023videocomposer, chen2023videocrafter1, videoworldsimulators2024, blattmann2023stable, girdhar2023emu} still exhibit randomness and face challenges in generating high-quality videos with controllability, which hinders the practical application of AIGC-based video generation methods.

\paragraph{Motion Control in Videos.} 
Recently, some studies have introduced additional control signals, such as trajectories~\citep{yin2023dragnuwa, wu2024draganything, wang2023motionctrl}, camera parameters~\citep{wang2023motionctrl, he2024cameractrl}, and bounding boxes~\citep{wang2024boximator}, to control visual elements in videos, \textit{i.e.}, camera transitions and object movements, thus achieving interactive video asserts generation. However, they lack the capability to precisely and finely manipulate visual elements, especially when it comes to object movements~\citep{yin2023dragnuwa, wang2023motionctrl}. In this paper, we meticulously design a training strategy that utilizes existing data to achieve flexible and precise motion separation and control.

\section{Preliminary}
\subsection{Conditional Video Diffusion Model}
Formally, diffusion models consists of a {forward process} and a {reverse  process}~\citep{sohl2015deep, ho2020denoising, song2020score}. The \emph{forward process} is defined as a Markov chain that progressively adds distinct levels of Gaussian noise to the signal $\x_0$ over a series of timesteps $t \in [0,T]$, until the $\x_0$ is completely corrupted to $\x_T\sim \mathcal{N}(\bm{0}, \I)$: 
\begin{align}
\label{eq:forward}
q(\x_{t}\mymid \x_{t-1}) = \mathcal{N}(\sqrt{\alpha_t}\x_{t-1}, (1-\alpha_t)\I),
\quad\text{and}\quad
q(\x_{t}\mymid \x_{0}) = \mathcal{N}(\sqrt{\widebar{\alpha}_t}\x_{0}, (1-\widebar{\alpha}_t)\I),
\end{align}
Here we consider the variance-preserving setting~\citep{score-sde} with $0 < \alpha_t < 1$ and $\widebar{\alpha}_t = \prod_{i = 1}^t \alpha_i$ where $\alpha_t$ is a decreasing sequence. The \emph{reverse process} is a parameter-containing process designed to iteratively denoise the corrupted sequence $\x_{T}$:
\begin{align}
p(\x_{t-1}\mymid \x_{t}) = \mathcal{N}(\bm{\mu}_t(\x_{t}), \sigma_t^2\I).
\end{align}
The mean and variance of the reverse process can be defined as:
\begin{align}
\bm{\mu}_t(\x_{t}, \x_{0}) &= \frac{\sqrt{\alpha_t}(1-\widebar{\alpha}_{t-1})}{1-\widebar{\alpha}_t}\x_{t} + \frac{\sqrt{\widebar{\alpha}_{t-1}}(1-\alpha_t)}{1-\widebar{\alpha}_{t}}\x_0,
\\
\sigma_t^2 &= (1-\alpha_t)\bigg(\frac{1-\widebar{\alpha}_{t-1}}{1-\widebar{\alpha}_t}\bigg).
\end{align}
Here we consider $\sigma_t^2$ is an untrained time dependent constants~\citep{ho2020denoising,im-ddpm}, and $\x_0$ can be reparameterized using Eq.~\ref{eq:forward} and estimated using $v$-prediction~\citep{prog-distill} or $\epsilon$-prediction techniques~\citep{ho2020denoising}.

Given an input condition $\c$, the goal of the conditional video diffusion model is to sample a video sequence $\x_0=\{\x_0^1,\x_0^2,\cdots,\x_0^L\}$ with $L$ frames from the conditional probability distribution $p(\x_0|\c)$. Specifically, $\bm{\mu}_\theta(\x_t, t, c)$ can be calculated using the $\epsilon$-prediction:
\begin{align}
\bm{\mu}_\theta(\x_t, t, \c) = \frac{1}{\sqrt{\alpha_t}}\bigg(\x_{t} - \frac{1-\alpha_t}{\sqrt{1-\widebar{\alpha}_t}}{\bm \epsilon}_\theta(\x_{t}, t, \c)\bigg),
\end{align}
where $\epsilon_\theta$ is a denoising UNet network. In this case, the $\epsilon_\theta$   is optimized via denoising score matching~\citep{song2019generative}:
\begin{align}
\min_{\theta} \mathbb{E}_{(\x_{0}, \c) \sim q(\x_{0}, \c), {\bm \epsilon} \sim \mathcal{N}(\bm{0}, \I), t}\big[\|{\bm \epsilon} - {\bm \epsilon}_{\theta}(\sqrt{\overline{\alpha}_t}\x_{0} + \sqrt{1-\overline{\alpha}_t}{\bm \epsilon}, t, \c)\|_2^2\big].
\end{align}
\subsection{Low-Rank Adaptation}
Low-Rank Adaptation (LoRA)~\citep{lora} is a parameter-efficient tuning approach used to accelerate model fine-tuning on incoming data, which can prevent catastrophic forgetting~\citep{ren2024analyzing}. Unlike training the entire model, LoRA adds a pair of rank-decomposition matrices to the linear layer weights, which optimizes only the newly introduced parameters and ensures that the other parameters are fixed. Mathematically, the new weights $W'\in\mathbb{R}^{m\times n}$ can be defined as:
\begin{align}
W'= W + \Delta W=W+AB^T,
\end{align}
where $A\in \mathbb{R}^{m \times r}$ and $B\in \mathbb{R}^{n \times r}$ are a pair of learnable matrices and $r \ll \min(m,n)$ is the rank to reduce the cost of fine-tuning.

\section{Experimental Details}

\subsection{Implementation details.} 
We use Animatediff v3~\citep{guo2023animatediff} combined with RGB SparseCtrl~\citep{guo2023sparsectrl} as our base model for image-to-video generation. We train only the motion ControlNet while keeping the UNet backbone weights frozen. The motion ControlNet is trained on our cultivated sampled 16-frame video sequences with a resolution of $384 \times 256$ (Section~\ref{Sec:3.2}). 
Both camera LoRA and object LoRA is optimized with Adam~\citep{kingma2014adam} on 8 NVIDIA Tesla V100 GPUs for a week with a batch size of 64 and a learning rate of $1\times10^{-4}$. We initially train the motion ControlNet using mixed data. 
Subsequently, we utilize camera-only data and mixed data to extract the camera LoRA and object LoRA weights respectively. To facilitate user input, we follow a strategy of training on dense trajectories first, and then fine-tuning the model on sparse trajectories. During the inference phase, we use 25 steps of DDIM sampler~\citep{song2020denoising}. Unless otherwise noted, the scale of classifier-free guidance~\citep{ho2022classifier} is set to 8.5.

\subsection{Evaluation metrics.}
To thoroughly evaluate the effectiveness of our method, we following MotionCtrl~\citep{wang2023motionctrl} to assessed two types of metrics: 1) Video content quality evaluation. We employ Fréchet Inception Distance (FID)\citep{fid}, Fréchet Video Distance ({FVD})\citep{fvd} to measure the visual quality and temporal coherenceand. The reference videos of FID and FVD are 5000 videos randomly selected from WebVid~\citep{bain2021frozen}. 
2) Video motion quality evaluation. The Euclidean distance between the predicted and ground truth trajectories, i.e., CamMC and ObjMC, is used to evaluate the motion control. Unlike MotionCtrl, which uses particleSFM~\citep{zhao2022particlesfm} to estimate the camera poses of the predicted video for calculating CamMC, we directly extract pixel-level movement trajectories to compute CamMC similar to ObjMC.
\begin{figure}
\centering
\includegraphics[width=0.9\textwidth]{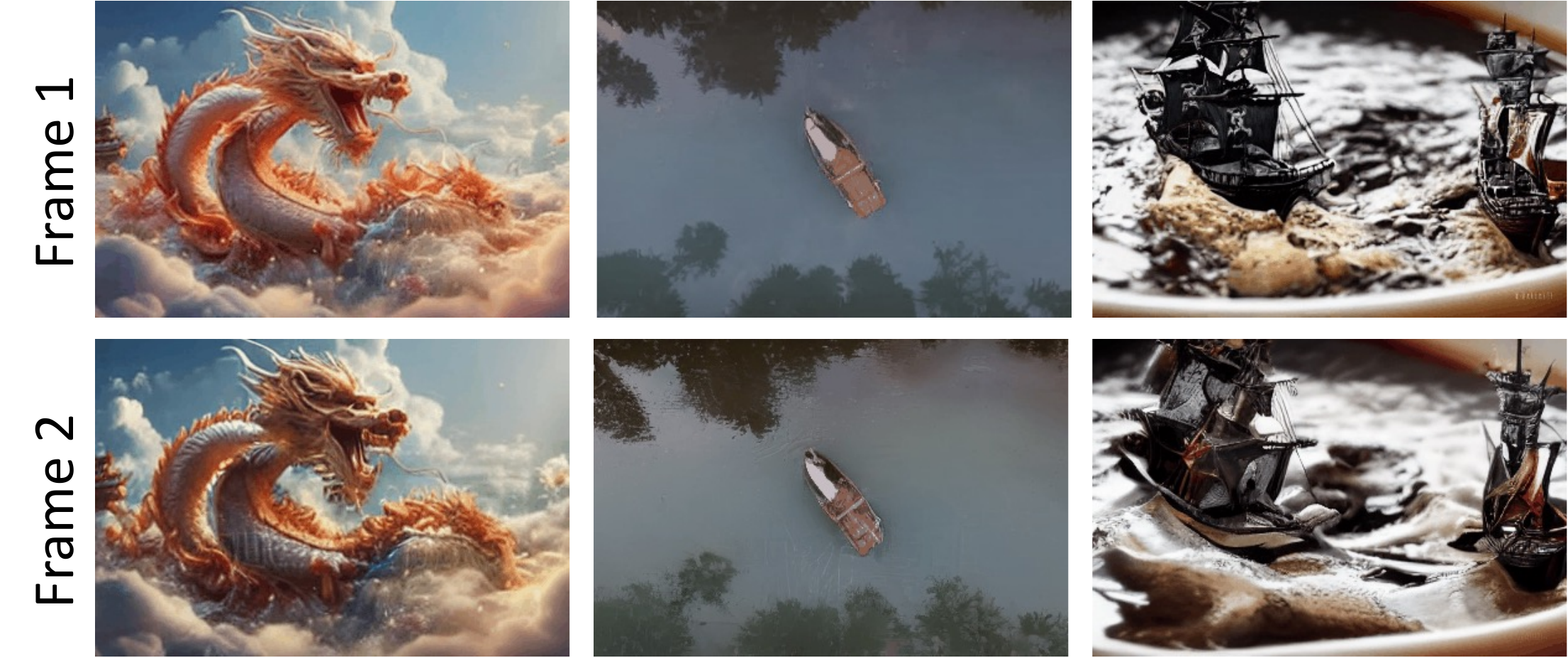}
\caption{Inherent video content inconsistency between the first frame and subsequent frames in the base model. \label{fig:fail1}}
\end{figure}

\section{Limitations}
Despite our model can faithfully produce motion information based on user-input trajectories, the generated quality of content is constrained by the base model. For example, as shown in Fig.~\ref{fig:fail1}, we observe that although Animatediff~\citep{guo2023animatediff} with image SparseCtrl~\citep{guo2023sparsectrl} imposes strong constraints on the first frame, subsequent frames exhibit some inconsistencies in color and detail compared to the first frame. One possible solution is to concatenate noisy image latents to the input noise in addition to using the image conditioning injection mechanism, similar to SVD~\citep{blattmann2023stable} and DynamiCrafter~\citep{xing2023dynamicrafter}. 

Another limitation is that despite text and image prompts generally complementing each other in most scenarios during the video generation process, if they convey different meanings, the quality of the output may be compromised.

\end{document}